\documentclass[lettersize,journal]{IEEEtran}
\IEEEoverridecommandlockouts

\usepackage{alphabeta}
\usepackage{cite}
\usepackage{mathtools,amsmath,amssymb,amsfonts,amsthm}
\usepackage{algorithmic}
\usepackage{graphicx}
\usepackage{textcomp}
\usepackage[dvipsnames]{xcolor}
\usepackage{pgfplots}
\pgfplotsset{compat=1.18}
\pgfplotsset{clean/.style={axis lines*=left,
        axis on top=true,
        axis x line shift=0.0em,
        axis y line shift=0.5em,
        every tick/.style={black, thick},
        axis line style = ultra thick,
        tick align=outside,
        clip=false,
        major tick length=4pt}}
\usepackage{cleveref}
\usepackage{tabularx}
\usepackage{multirow}
\usepackage{booktabs}

\definecolor{tolblue}{rgb}{0.2667,0.4667,0.6667}
\definecolor{tolred}{rgb}{0.9333,0.4000,0.4667}
\definecolor{tolyellow}{rgb}{0.8000,0.7333,0.2667}
\definecolor{tolcyan}{rgb}{0.4000,0.8000,0.9333}
\definecolor{tolpurple}{rgb}{0.6667,0.2000,0.4667}
\definecolor{tolgreen}{rgb}{0.1333,0.5333,0.2000}

\usepackage[listings,skins,breakable]{tcolorbox}
{%
\end{tcolorbox}\endgroup}

{%
\end{tcolorbox}\endgroup}

\DeclareMathOperator*{\argmin}{arg\,min}

\def\BibTeX{{\rm B\kern-.05em{\sc i\kern-.025em b}\kern-.08em
    T\kern-.1667em\lower.7ex\hbox{E}\kern-.125emX}}
\begin{document}

\title{Pose Tracking with a Foundation Pose Model and an Ensemble Directional Kalman Filter\thanks{Identify applicable funding agency here. If none, delete this.}
}


\author{Tianlu Lu, Asif Sijan, Thomas Noh, Huaijin Chen, Andrey A. Popov.
\thanks{Manuscript received \today, This work was supported by FUNDING.}%
\thanks{T. Lu, A. Sijan, H. Chen, and A.A. Popov are with the Information and Computer Sciences Department at the University of Hawaii at Manoa. T. Noh is with the John A. Burns School of Medicine at the University of Hawaii at Manoa.}}

\maketitle

\begin{abstract}
This paper introduces the ensemble directional Kalman filter (EnDKF), an ensemble-based Kalman filtering approach for pose tracking that jointly estimates an object's position and attitude using ideas from directional statistics. 
The EnDKF integrates a unit-quaternion attitude representation to move beyond canonical Kalman filter mean and covariance assumptions that poorly capture directional uncertainty.
Experiments on a synthetic constant-velocity constant-angular-velocity system and a digital-twin head-tracking scenario using the FoundationPose algorithm demonstrate a significant reduction in error as opposed to merely using measurements.
\end{abstract}


\section{Introduction}

This work brings together concepts from two traditionally distinct areas: estimation and data assimilation~\cite{jazwinski2007stochastic,asch2016data,reich2015probabilistic} on the one hand, and computer vision on the other. 
In computer vision and deep learning, pose estimation~\cite{zheng2023deep} typically denotes the computation of an object’s position and attitude from sensory data, without associated uncertainty quantification and usually without the fusion of prior information. Within the estimation community, such computations are more naturally described as sensing or detection, while estimation refers to the sequential fusion of prior uncertainty with these detections. To avoid ambiguity, we use the term \textit{pose tracking} to denote the sequential estimation of uncertainty over an object’s pose. The objective of this work is to combine a computer-vision-based pose detector with a novel ensemble Kalman filtering methodology for pose tracking.

We employ the state-of-the-art computer vision-based pose detection algorithm FoundationPose~\cite{wen2024foundationpose}, which produces a sequence of measurements of an object's position and attitude. 
For a single measurement, FoundationPose generates global pose hypotheses around an initial detection and iteratively refines them by exploiting color and depth information. 
The pose state consists of position and attitude, where position is naturally represented in Euclidean space, and attitude requires a representation on a nonlinear manifold.
To this end, we adopt a unit quaternion representation~\cite{shuster1993survey} of attitude so that it would fall within the framework of directional statistics.

The multiplicative Kalman filter (MKF)~\cite{zanetti2009multiplicative, zanetti2018fully} takes advantage the Kalman filter~\cite{kalman1960new} equations together with small-angle approximations to update attitude states. 
Its reliance on Euclidean approximations renders it poorly suited to settings in which uncertainty is more accurately captured by directional distributions whose behavior is not well summarized by mean and covariance alone. 
Related approaches~\cite{subrahmanya2025preserving} that enforce spherical or manifold constraints via iterative projection or constrained optimization also ultimately depend on mean and covariance approximations, and thus inherit similar limitations.
In contrast, directional statistics~\cite{ley2017modern,pewsey2021recent,kurz2019directional} focus on inherently directional data, including attitude and other quantities naturally modeled on the $(n-1)$-sphere, $\mathbb{S}^{n-1}$, such as spherical word embeddings~\cite{meng2019spherical}. 
Filters designed specifically for such data have been proposed, including unscented-type filters~\cite{kurz2016unscented} and methods for nonlinear measurement models~\cite{gilitschenski2015non}, but their construction makes it difficult to handle joint uncertainty coupling directional and Euclidean components, as required for full pose estimation, with which this work is primarily concerned.

Directional statistics~\cite{ley2017modern,pewsey2021recent,kurz2019directional} deal with data that are purely directional in nature, most importantly for this work attitude, but also any data that can be embedded onto $\mathbb{S}^{n-1}$, such as directional word embeddings~\cite{meng2019spherical}.
Filters that deal purely with directional data have been developed, such as an unscented filter~\cite{kurz2016unscented}, and a for the nonlinear measurement setting~\cite{gilitschenski2015non}.
The way in which these filters are constructed would have difficulty dealing with dependence between directional and non-directional uncertainty, such as what would be required for full pose estimation.

This reveals a gap in the literature: the absence of a linear, Kalman-style filter that can jointly incorporate uncertainty in both position and attitude, including their co-dependencies (like covariance), without relying on restrictive mean and covariance assumptions in the directional components.
To address this gap, we derive a more fundamental form of the Kalman filter that operates on general manifolds, including spherical manifolds, and render it computationally tractable through an ensemble approximation in the spirit of the ensemble Kalman filter (EnKF)~\cite{Evensen_1994,Burgers_1998_EnKF}. 
We refer to the resulting method as the ensemble directional Kalman filter (EnDKF). 
We present this EnDKF in an alternative formulation of the perturbed-observations EnKF that corrects several biased approximations.
Finally, we extend the EnDKF to handle states comprising velocity, position, angular velocity, and attitude, thereby enabling estimation under a constant-velocity, constant-angular-velocity model, and demonstrate its performance on both a synthetic constant-velocity experiment and a digital twin head-tracking scenario.


\section{Background and Motivation}
\label{sec:background}

We first provide an unorthodox, but known, derivation of the Kalman filter that will enable us to re-derive and extend the ensemble Kalman filter for directional data.

Suppose that we known information from the prior represented by the random variable $X^-$. Suppose also that we are given a measurement of the truth,
\begin{equation}
    y = h(X^\text{true}) + \eta,
\end{equation}
where $h$ is some nonlinear measurement function and $\eta$ is unbiased additive error with known covariance $\operatorname{Cov}[Y, Y]$. It is important to note here that $\eta$ is not necessarily Gaussian.

We can define the family of linear estimators,
\begin{equation}
        X^\sim(G) = X^- + G\left( y - h(X^-)\right),
\end{equation}
parameterized by the linear gain matrix $G$.
The output $X^\sim(G)$ for some arbitrary gain is an approximation to the posterior.
Our goal is to find $G$ that incorporates the information from the measurement and the prior in a way that somehow minimizes our uncertainty. 
For instance if $G=0$ we do not incorporate measurement information at all. If $h = \operatorname{id}$, the identity function, and $G=I$, the identity matrix, then we fully ignore the prior and only incorporate measurement information.
All choices of $G$ lead to some approximation to the Bayesian posterior, with our goal being to pick the `correct' one.

If we wish to minimize our uncertainty in the posterior, a common metric that can be used is the trace-generalized variance,
\begin{equation}\label{eq:real-variance}
    \operatorname{Var}^{\mathbb{R}^n}\left(X\right) = \operatorname{tr}\left[\operatorname{Cov}\left(X\right)\right],
\end{equation}
where by $R^n$ we denote that this is the natural generalized variance that makes sense for Euclidean space.

We are then ready to define the `best' estimator according to our metric~\cref{eq:real-variance},
\begin{equation}\label{eq:Kalman-filter}
\begin{gathered}
    X^+ = X^\sim(K),\\
    K = \argmin_{G}\operatorname{Var}\left[ X^\sim(G)\right],
\end{gathered}
\end{equation}
where $K$ is called the Kalman gain, and the equation above we call the Kalman filter.
We stop at this point in the derivation \textit{without} explicitly solving for the Kalman gain, as this allows us to generalize the Ensemble Kalman filter later in this work.
It is very important to note that this derivation has not used any assumptions about distributions: this means that the Kalman filter does not rely on many a Gaussian assumption at all unlike other previous attempts at the problem~\cite{ge2025geometryextendedkalmanfilters}.
The posterior in~\cref{eq:Kalman-filter} is known as the best linear unbiased estimator.

If the prior has known finite mean and finite covariance,   the measurement operator does not cause the mean or covariance to become infinite, and the measurement has known finite covariance, then a closed form solution for the Kalman gain is known in this case,
\begin{equation}\label{eq:Kalman-gain-exact}
    K = \operatorname{Cov}[X, h(X)]\left(\operatorname{Cov}[h(X), h(X)] + \operatorname{Cov}[Y, Y]\right)^{-1},
\end{equation}
for which there are a multitude of ways of either keeping track of the covariances, or estimating the covariances. 
In general, it is intractable to compute~\cref{eq:Kalman-gain-exact} exactly, therefore in this work we focus on the ensemble Kalman filter (EnKF) type algorithms, derived in~\cite{Evensen_1994,Burgers_1998_EnKF}.

As some of the covariances in~\cref{eq:Kalman-gain-exact} are intractable to compute without linearizations, the traditional (perturbed observations) EnKF, focuses its efforts on the estimation of the covariances through ensemble covariances.
Given an ensemble of $N$ samples from the prior,
\begin{equation}
    E_{X^-} = \left[X^-_1, \dots X^-_N\right],
\end{equation}
the perturbed observations ensemble Kalman filter approximates the covariances as,
\begin{equation}\label{eq:ensemble-covariance}
\begin{aligned}
    \operatorname{Cov}[X, h(X)] &= (N-1)^{-1}E_{X^-} C h(E_{X^-})^T,\\
    \operatorname{Cov}[h(X), h(X)] &= (N-1)^{-1}h(E_{X^-}) C h(E_{X^-})^T,\\
    C &= I_N - N^{-1} 1_N 1_N^T,
\end{aligned}
\end{equation}
where $h$ is applied ensemble-wise, $I_N$ is the identity matrix of size $N$, and $1_N$ is a column vector of ones.

The perturbed observations EnKF is then defined on the family of linear estimators,
\begin{equation}\label{eq:linear-ensemble-estimator}
    E_{X^\sim}(G,\omega) = E_{X^-} + G\left( E_{Y}(\omega) - h(E_{X^-})\right),
\end{equation}
where $\omega$ is a random variable that determines the ensemble of perturbed observations, $E_Y(\omega)$, which are defined as,
\begin{equation}\label{eq:Rn-perturbed}
    E_Y(\omega) = \left[ y - \eta_1(\omega), \dots, y - \eta_N(\omega) \right],
\end{equation}
where each $\eta_i(\omega)$ comes either from the same distribution as $\eta$ or from some other unbiased distribution with the same known covariance (usually a Gaussian).

Approximating the exact form of the Kalman gain in~\cref{eq:Kalman-gain-exact} through the ensemble covariance approximation in~\cref{eq:ensemble-covariance}, and obtaining some $\widetilde{K}$, the perturbed observations EnKF posterior is defined as,
\begin{equation}\label{eq:POEnKF}
    E_{X^+} =E_{X^\sim}(\widetilde{K},\omega),
\end{equation}
again for one single choice of $\omega$. 

We identify the following problems with this approach that this paper all aims to solve:
\begin{enumerate}
    \item the ensemble Kalman gain approximation obtained through the covariance approximation in~\cref{eq:ensemble-covariance} is biased, as the quotient of two unbiased random variables is a biased estimator,
    \item the EnKF posterior in~\cref{eq:POEnKF} is defined only for a single choice of $\omega$, and
    \item the EnKF does not generalize to dealing with data that live on non-Euclidean manifolds, such as those that occur in directional statistics.
\end{enumerate}
All three of these problems are addressed in the next section.

\section{Reformulating the EnKF}
\label{sec:enkf}

The goal of this section is to reformulate the ensemble Kalman filter such that it addresses all three problems identified in the previous section.

We again start with the perturbed observations family of linear estimators from~\cref{eq:linear-ensemble-estimator},
where for some choice of $G$ we can minimize our uncertainty.
We first define the trace generalized variance~\cref{eq:real-variance} of the ensemble $E_X$, as,
\begin{equation}
    \operatorname{Var}^{\mathbb{R}_n}(E_X) = \sum_{i,j} \left[(\sqrt{N - 1})^{-1} E_X C)^{\circ 2}\right]_{i,j},
\end{equation}
where $C$ is the centering matrix from~\cref{eq:ensemble-covariance}, and the squaring by $\circ 2$ is performed element-wise.

We are now ready to tackle the first two problems identified in the previous section: the fact that the Kalman gain is a biased estimator of the solution to the optimization problem defined in~\cref{eq:Kalman-filter}, and that the posterior only depends on one value of $\omega$.
We discard the ensemble Kalman gain formulation defined by the equations~\cref{eq:ensemble-covariance}, and reformulate the Kalman gain as,
\begin{equation}\label{eq:new-Kalman-gain}
    K = \argmin_{G}\mathbb{E}_\omega\left[\operatorname{Var}^{\mathbb{R}^n}\left[E_{X^\sim}(G, \omega) \right] \right],
\end{equation}
which is now a stochastic optimization problem. In essence, we wish to pick the Kalman gain to be the gain matrix that minimizes the ensemble trace generalized variance of the linear ensemble estimator in~\cref{eq:linear-ensemble-estimator} over all possible realizations of $\omega$. This formulation of the ensemble Kalman filter is a different stochastic interpretation of the Kalman filter equations in~\cref{eq:Kalman-filter}.

By making one more choice of an $\omega^*$, which is no different than all the other choices of $\omega$, we can define the new ensemble Kalman filter posterior as,
\begin{equation}
    E_{X^+} = E_{X^\sim}(K, \omega^*),
\end{equation}
where the new posterior ensemble depends on the new Kalman gain in~\cref{eq:new-Kalman-gain} which i) does not rely on the ensemble covariances~\cref{eq:ensemble-covariance} and does not rely on only one draw of $\omega$ because of the stochastic optimization problem.

We next move on to the final problem that we wish to address with the Kalman filter: that it does not generalize to non-Euclidean manifolds.
We first define the exponential, logarithmic, and projection maps for Euclidean space.
We begin by defining the exponential map for Euclidean space, which defines the geodesic along the direction $V$ starting from the point $X_1$, as,
\begin{equation}\label{eq:Rn-exp}
    \operatorname{exp}_{X_1}^{\mathbb{R}^n}(V) = X_1 + V
\end{equation}
which is simply defined by vector addition.
The corresponding logarithmic map takes two vectors $X_1$ and $X_2$ and asks, what vector $V$ would, as an input to the exponential map from the point $X_1$, result in the vector $X_2$, and is defined as,
\begin{equation}\label{eq:Rn-log}
     \operatorname{log}_{X_1}^{\mathbb{R}^n}(X_2) = X_2 - X_1,
\end{equation}
which again is simply vector subtraction.
We note that these two operations have the properties:
\begin{equation}\label{eq:Rn-P}
\begin{gathered}
    \operatorname{exp}_{X_1}^{\mathbb{R}^n}\left(0\right) = X_1,\\
    \operatorname{exp}_{X_1}^{\mathbb{R}^n}\left(\operatorname{log}_{X_1}^{\mathbb{R}^n}(X_2)\right) = X_2,
\end{gathered}
\end{equation}
which fully defines the geodesic in Euclidean space from the point $X_1$ to the point $X_2$.
Finally we define the projection operator that maps vectors to the tangent space,
\begin{equation}
    P^{\mathbb{R}^n}_{X^-}\left(V\right) = V,
\end{equation}
which for Euclidean space is trivially the identity.

We are now ready to redefine the ensemble Kalman filter for Euclidean space in terms of the new 
\begin{equation}\label{eq:Rn-EnKF}
\begin{gathered}
    E_{X^\sim}(G,\omega) = \operatorname{exp}_{E_{X^-}}^{\mathbb{R}^n}\left(P^{\mathbb{R}^n}_{X^-}\left[ G \operatorname{log}_{h(E_{X^-})}^{\mathbb{R}^n}(E_Y(\omega))\right]\right),\\
    E_{X^+} = E_{X^\sim}(K, \omega^*),\\
    K = \argmin_{G}\mathbb{E}_\omega\left[\operatorname{Var}^{\mathbb{R}^n}\left( E_{X^\sim}(G, \omega) \right) \right],
\end{gathered}
\end{equation}
where the first equation defines the general family of linear maps in Euclidean space,
the second equation again defines the posterior in terms of the Kalman gain $K$ which itself is again defined by the third equation.

This form of the filter has several interesting properties. First, the filter does not directly rely on the Euclidean distance between the actual measurements and our prior ensemble in measurement space, but instead relies on the log-map between the two. Second, the filter still relies on a linear control parameter, in terms of the gain matrix $G$. Third, the projection of the multiplication of $G$ by the log map ensures that the input to the exponential map is valid, and fourth, the filter is constructed to be the best linear estimator in this class of filters. 
Finally, this form of the filter is generalizable to any manifold for which the exponential map, logarithmic map, projection operators, and generalized variance are known and can be feasibly computed over ensembles.

\subsection{Stochastic Optimization}

The stochastic optimization in~\cref{eq:Rn-EnKF} likely does not have a tractable closed form solution in the case of a finite ensemble. This means that a numerical approximation must be utilized. 
In this work we make use of the Adam~\cite{kingma2014adam} stochastic optimization technique for its ease of implementation and speed, but not for any theoretical properties.
It is of independent interest to find custom tailored stochastic optimization techniques for this class of equation.

As an initial condition $G_0$ we take the naive ensemble Kalman gain, computed through~\cref{eq:ensemble-covariance} and~\cref{eq:Kalman-gain-exact}, where the covariance $\operatorname{Cov}[Y,Y]$ can be numerically computed through the use the perturbed observations.
We initially take one realization of the noise $\omega_0$ and compute the gradient of the variance, and set,
\begin{equation}
    M_0 = 0,\quad  V_0 = \left[\nabla_G \operatorname{Var}^{\mathbb{R}^n}\left( E_{X^\sim}(G_0, \omega_0) \right)\right]^{\circ 2},
\end{equation}
which are the momentum and velocity terms, and update,
\begin{equation}
    G_1 = G_0 - \alpha \frac{M_0}{\sqrt{V_0 + \epsilon}},
\end{equation}
where the division is element-wise, $\epsilon$ is a numerical stability parameter defined later, and $\alpha$ is the step-size/learning-rate.

For all subsequent steps $i = 1, \dots$, we take a realization of the noise $\omega_i$, and compute,
\begin{equation}\label{eq:adam}
    \begin{aligned}
        D_i &= \nabla_G \operatorname{Var}^{\mathbb{R}^n}\left( E_{X^\sim}(G_i, \omega_i) \right)\\
        M_i &= \beta_1 M{i-1} + (1 - \beta_1) D_i,\\
        V_i &= \beta_2 V{i-1} + (1 - \beta_2) D_i^{\circ 2},\\
        G_i &= G_{i-1} - \alpha \frac{M_i}{\sqrt{V_i + \epsilon}},
    \end{aligned}
\end{equation}
where $\beta_1$ and $\beta_2$ are parameters that control the rate at which the momentum and velocity change.

As we are using Adam for a stochastic optimization problem where it is possible to generate infinite data, we do not treat notions of epoch, or batch, as these notions are not directly applicable to this problem.
As we aim for speed in the algorithm, We take $\beta_1$ to be standard, but $\beta_2$ to be lower than the standard, 
\begin{equation}
    \beta_1 = 0.9,\quad \beta_2 = 0.95,\quad \epsilon = 10^{-8},
\end{equation}
to ensure that we can converge to a useful solution in a rapid manner.

\section{Directional EnKF}
\label{eq:directional-enkf}

We are now ready to generalize the equations in~\cref{eq:Rn-EnKF} to a non-Euclidean manifold.
As in this work we are concerned with directional data, we focus on the sphere with attribute dimension $n$.
As before we need to define the exponential map, the logarithmic map, the projection operator, and a measure of variance on the sphere.

For the sphere with attribute dimension $n$, $\mathbb{S}^{n-1}$, we can similarly define the exponential map~\cite{manoptjl_sphere},
\begin{equation}\label{eq:Snm1-exp}
    \operatorname{exp}_{X_1}^{\mathbb{S}^{n-1}}(V) = \cos\left(\lVert V\rVert \right) X_1 + \sin\left(\lVert V\rVert\right)\frac{V}{\lVert V\rVert},
\end{equation}
which defines the geodesic starting at point $X_1$ in the direction $V$.
The inverse of this, the logarithmic map is similarly defined as,
\begin{equation}\label{eq:Snm1-log}
     \operatorname{log}_{X_1}^{\mathbb{S}^{n-1}}(X_2) = \operatorname{acos}\left( X_1^T X_2 \right)\frac{P_{X_1}(X_2 - X_1)}{\lVert P_{X_1}(X_2 - X_1) \rVert},
\end{equation}
where the projection operator is defined as,
\begin{equation}\label{eq:Snm1-P}
    P^{\mathbb{S}^{n-1}}_{X_1}(X_2) = X_2 - X_1 (X_1^T X_2),
\end{equation}
that combined define how `linear' operations on the sphere are performed. The action of the projection operator is to make sure that any input $X_2$ is forced to lie on the tangent bundle associated with $X_1$.

We then discuss some possible non-additive errors for the directional statistics case. The isotropic diagonal Gaussian distribution distribution restricted to $\mathbb{S}^{n-1}$ is the Fisher-von~Mises-Langevin distribution,
\begin{equation}
    \mathcal{FML}(x; \mu, \kappa) \propto e^{\kappa x^T \mu}, 
\end{equation}
where $\mu$ is known as the mean direction and has Euclidean norm one, and $\kappa$ is what is known as the concentration and corresponds to the diagonal element of the precision matrix of the underlying isotropic diagonal Gaussian distribution.

The error in observing directional data can be modeled as,
\begin{equation}
    y \sim \mathcal{FML}(X^\text{true}, \rho),
\end{equation}
where $X^\text{true}$ is the (true) mean direction of the data, and $\rho$ is what is the concentration parameter.
An ensemble of perturbed observations can be defined for this case as,
\begin{equation}\label{eq:Snm1-perturbed}
\begin{gathered}
    E_Y(\omega) = \left[ y_1(\omega), \dots, y_N(\omega) \right],\\
    y_i(\omega) \sim \mathcal{FML}(y, \rho),
\end{gathered}
\end{equation}
where each realization $y_i$ is determined by $\omega$ but pair-wise independent.

In order to complete the set of operations, we need to  find a metric that describes the variance on the sphere. While it is possible to make use of the concentration, estimating the concentration parameter is difficult, and no unbiased estimator is possible for Fisher-von~Mises-Langevin distribution (which is proven in unpublished work).
We make the observation that the concentration parameter is the spherical analogue of the inverse of the covariance, and grows as the norm of the mean of a random variable increases. Furthermore the maximum norm of the mean of a random variable is one in the degenerate case. Thus, we make use of the following generalized variance analogue,
\begin{equation}\label{eq:Sphere-Variance}
    \operatorname{Var}^{\mathbb{S}^{n-1}}(E_X) = 1 - \left\lVert N^{-1} \sum_{i=1}^N [E_X]_i \right\rVert^2,
\end{equation}
which is a modification of the circular variance in~\cite{allen1991automated, berens2009circstat}, with the square of the norm used to make sure the units of the variance are correct.
As the mean estimator inside the norm is bounded from above by one for the case of the sphere, the norm squared is always less than one, thus one minus this value is minimized when all the values of the ensemble are the same, thus have no variance.

Combining all this together, we can modify the new ensemble Kalman filter equations in~\cref{eq:Rn-EnKF} to create its directional analogue:
\begin{equation}\label{eq:Snm1-EnKF}
\begin{gathered}
    E_{X^\sim}(G,\omega) = \operatorname{exp}_{E_{X^-}}^{\mathbb{S}^{n-1}}\left(P^{\mathbb{S}^{n-1}}_{X^-}\left[ G \operatorname{log}_{h(E_{X^-})}^{\mathbb{S}^{n-1}}(E_Y(\omega))\right]\right),\\
    E_{X^+} = E_{X^\sim}(K, \omega^*),\\
    K = \argmin_{G}\mathbb{E}_\omega\left[\operatorname{Var}^{\mathbb{S}^{n-1}}\left( E_{X^\sim}(G, \omega) \right) \right],
\end{gathered}
\end{equation}
which again it is trivial to solve using a slightly modified Adam procedure in~\cref{eq:adam}.
We term this filter the ensemble directional Kalman filter (EnDKF).

The EnDKF equations~\cref{eq:Snm1-EnKF} require the use of the projection operator as it is possible that the application of some gain matrix $G$ would result in a violation of the 

\section{Constant Velocity and Rotation Model}
\label{sec:constant-velocity-EnKF}

While there are use-cases where a purely directional filter can be useful, this work focuses on determining position and attitude. We model the attitude as a unit quaternion (versor) meaning that our estimate of the attitude part of the pose is directional data living in $\mathbb{S}^3$.

As the simplest proof of concept we  analyze the constant velocity constant angular velocity model. Consider the state of an object as represented by i) a vector $v$ in $\mathbb{R}^3$ representing the objects current velocity, ii) a vector $p$ in $\mathbb{R}^3$ representing the current position of the object, iii) a unit quaternion $\dot\omega$ in $\mathbb{S}^3$ representing the current angular velocity of the object, and iv) a unit quaternion $\omega$ in $\mathbb{S}^3$ representing the current attitude of the object.
We can therefore represent our state $x$ as the following column vector,
\begin{equation}\label{eq:constant-velocity-state}
    x = \begin{bmatrix} v & p & \dot{\omega} & \omega\end{bmatrix}^T,
\end{equation}
where the column vector represents a state in $\mathbb{R}^3\times \mathbb{R}^3\times \mathbb{S}^3\times \mathbb{S}^3$, and can be embedded into the space $\mathbb{R}^{14}$.

We can introduce process noise into the model by perturbing the state by unbiased Gaussian noise. For the velocity and position, we can define a velocity process noise covariance, $Q_v$, and position process noise covariance $Q_p$. For the directional data we take advantage of the Fisher-von~Mises-Langevin distribution and set a corresponding angular velocity process noise concentration of $\text{\Qoppa}_{\dot\omega}$ and a attitude process noise concentration of $\text{\Qoppa}_\omega$.

The constant velocity model therefore propagates a state from time index $k$ to time index $k+1$, with some $dt$, as follows:
\begin{equation}\label{eq:constant-velocity-model}
\begin{aligned}
    v_{k+1} &\sim \mathcal{N}(v_k, dt Q_v),\\
    p_{k+1} &= \mathcal{N}(p_k + dt \cdot v_k, dt Q_p),\\
    \dot{\omega}_{k+1} &\sim \mathcal{FML}(\dot{\omega}_k, dt^{-1}\text{\Qoppa}_{\dot{\omega}}),\\
    \omega_{k+1} &\sim \mathcal{FML}(\dot\omega_k^{dt}\cdot \omega_k, dt^{-1}\text{\Qoppa}_{\omega}),
\end{aligned}
\end{equation}
where the angular velocity quaternion is raised to a scalar power in the canonical way, and the concentration in the Fisher-von~Mises-Langevin distribution is multiplied by the inverse of the time interval to keep in line with the concentration corresponding to the precision matrix.

For the rest of this work, for simplicity, we treat the time interval as a constant $dt=1\,\text{f}$, where $f$ is a frame.

While the model itself treats the velocity and angular velocity as constant, our uncertainty about those constants allows us to track the pose of objects that behave in unpredictable ways. In order to define this tracking we must modify our new Kalman filter formulation to work with this model.

We combine our new formulation of the ensemble Kalman filter~\cref{eq:Rn-EnKF}, and the directional ensemble Kalman filter~\cref{eq:Snm1-EnKF}, to work with the state defined in~\cref{eq:constant-velocity-state}.
We first define the exponential map,
\begin{equation}\label{eq:const-vel-exp}
    \operatorname{exp}^{C}_{X}(V) = \begin{bmatrix}
    \operatorname{exp}^{\mathbb{R}^3}_{X_v}(V_v)\\
    \operatorname{exp}^{\mathbb{R}^3}_{X_p}(V_p)\\
    \operatorname{exp}^{\mathbb{S}^3}_{X_{\dot{\omega}}}(V_{\dot{\omega}})\\
    \operatorname{exp}^{\mathbb{S}^3}_{X_{\omega}}(V_{\omega})
    \end{bmatrix}
\end{equation}
which applies the exponential map in Euclidean space~\cref{eq:Rn-exp} on the velocity $v$ and position $p$, and the exponential map in directional space~\cref{eq:Snm1-exp} on the angular velocity quaternion $\dot\omega$ and attitude quaternion $\omega$.
Next we define the logarithmic map the same exact way as in~\cref{eq:const-vel-exp},
%
which applies the logarithmic map in Euclidean space~\cref{eq:Rn-log} on the velocity $v$ and position $p$, and the logarithmic map in directional space~\cref{eq:Snm1-log} on the angular velocity quaternion $\dot\omega$ and attitude quaternion $\omega$.
Finally we define the projection map the same exact way as in~\cref{eq:const-vel-exp},
%
%
which applies the projection map in Euclidean space~\cref{eq:Rn-P} on the velocity $v$ and position $p$, and the projection map in directional space~\cref{eq:Snm1-P} on the angular velocity quaternion $\dot\omega$ and attitude quaternion $\omega$.

We then define the way in which the ensemble containing all our uncertainty can be split.
If $X$ represents our uncertainty about the total state~\cref{eq:constant-velocity-state}, then we can say that $X_v$ represents our uncertainty about the velocity, $X_p$ represents our uncertainty about position, $X_{\dot\omega}$ represents our uncertainty about the angular velocity and $X_\omega$ represents our uncertainty about the attitude.
Therefore, an ensemble

We next define the generalized variance,
\begin{equation}
\begin{multlined}
    \operatorname{Var}^C(E_C) = \operatorname{Var}^{\mathbb{R}^3}(E_{X_v}) + \operatorname{Var}^{\mathbb{R}^3}(E_{X_p}) + \\ \operatorname{Var}^{\mathbb{S}^{3}}(E_{X_{\dot{\omega}}}) + \operatorname{Var}^{\mathbb{S}^{3}}(E_{X_{\omega}})
\end{multlined}
\end{equation}
as the sum of the variance in position and velocity defined by their individual Euclidean trace-generalized variances~\cref{eq:real-variance} and the variance in angular velocity and attitude defined by their individual spherical variances~\cref{eq:Sphere-Variance}.

For the perturbed observations $E_Y(\omega)$, those depend on the observation operator $h$ and the distributions from which they come, but can again be some form of concatenation.

Finally, the generalized linear family of filters we define as
\begin{equation}\label{eq:EnDKF-C}
    \begin{gathered}
    E_{X^\sim}(G,\omega) = \operatorname{exp}_{E_{X^-}}^{C}\left(P^{C}_{X^-}\left[ G \operatorname{log}_{h(E_{X^-})}^{C}(E_Y(\omega))\right]\right),\\
    E_{X^+} = E_{X^\sim}(K, \omega^*),\\
    K = \argmin_{G}\mathbb{E}_\omega\left[\operatorname{Var}^{C}\left( E_{X^\sim}(G, \omega) \right) \right],
\end{gathered}
\end{equation}
where $G\in \mathbb{R}^{14\times 7}$ is a gain matrix, and $K\in \mathbb{R}^{14\times 7}$, is the generalized Kalman gain.

For the remained of this work we assume that we measure the position and the rotation but do not measure the velocity or the angular velocity,
\begin{equation}\label{eq:measurements}
    y = \begin{bmatrix}
        y_p \\ y_\omega
    \end{bmatrix},
\end{equation}
with some assumed known position covariance $R \in \mathbb{R}^{3\times 3}$ and some known attitude concentration $\rho\in\mathbb{R}$.
Our model of the measurement operator therefore looks like,
\begin{equation}\label{eq:measurement-pos-att}
    h(E_X) = \begin{bmatrix}
         E_{X_p}\\
         E_{X_\omega}
    \end{bmatrix},
\end{equation}
taking the ensemble and extracting out the position and attitude.

\subsection{Constant Bias Model}

In the realistic head-tracking application considered in this work, the measurement process is subject to systematic bias.
When such bias in the measurement function is known to exist, we can extend the state-space model to include constant bias terms.
Doing so will allow the EnDKF to jointly estimate both the true head pose and the associated bias, thereby improving the overall accuracy over a naive measurements-only approach.

We can model two new variables to represent the bias terms. The variable $b\in\mathbb{R}^3$, with corresponding exponential~\cref{eq:Rn-exp}, logarithmic~\cref{eq:Rn-log}, and projection maps~\cref{eq:Rn-P}, represents the linear bias in our position measurements. 
The variable $\beta\in\mathbb{S}^3$, with corresponding exponential~\cref{eq:Snm1-exp}, logarithmic~\cref{eq:Snm1-log}, and projection maps~\cref{eq:Snm1-P}, represents the multiplicative bias in our attitude measurements.
Therefore we can augment the measurement operator in~\cref{eq:measurement-pos-att}, as,
\begin{equation}\label{eq:measurement-pos-att-bias}
    h(E_X) = \begin{bmatrix}
         E_{X_p} + E_{X_b}\\
         E_{X_\beta}\cdot E_{X_\omega}
    \end{bmatrix},
\end{equation}
where the position bias is linear and the attitude bias is multiplicative.
The biases can be propagated through time in addition to the constant velocity model~\cref{eq:constant-velocity-model}, as,
\begin{equation}\label{eq:constant-bias-model}
\begin{aligned}
    b_{k+1} &\sim \mathcal{N}(b_k, dt Q_b),\\
    \beta_{k+1} &\sim \mathcal{FML}(\beta_k, dt^{-1}\text{\Qoppa}_{\beta}),
\end{aligned}
\end{equation}
where $Q_b\in\mathbb{R}^{3\times 3}$ is the bias covariance and $\text{\Qoppa}_{\beta} \in\mathbb{R}$ is the bias concentration.

\section{Numerical Experiments}
\label{sec:numerics}

\begin{figure}
    \centering
\includegraphics[width=0.95\linewidth]{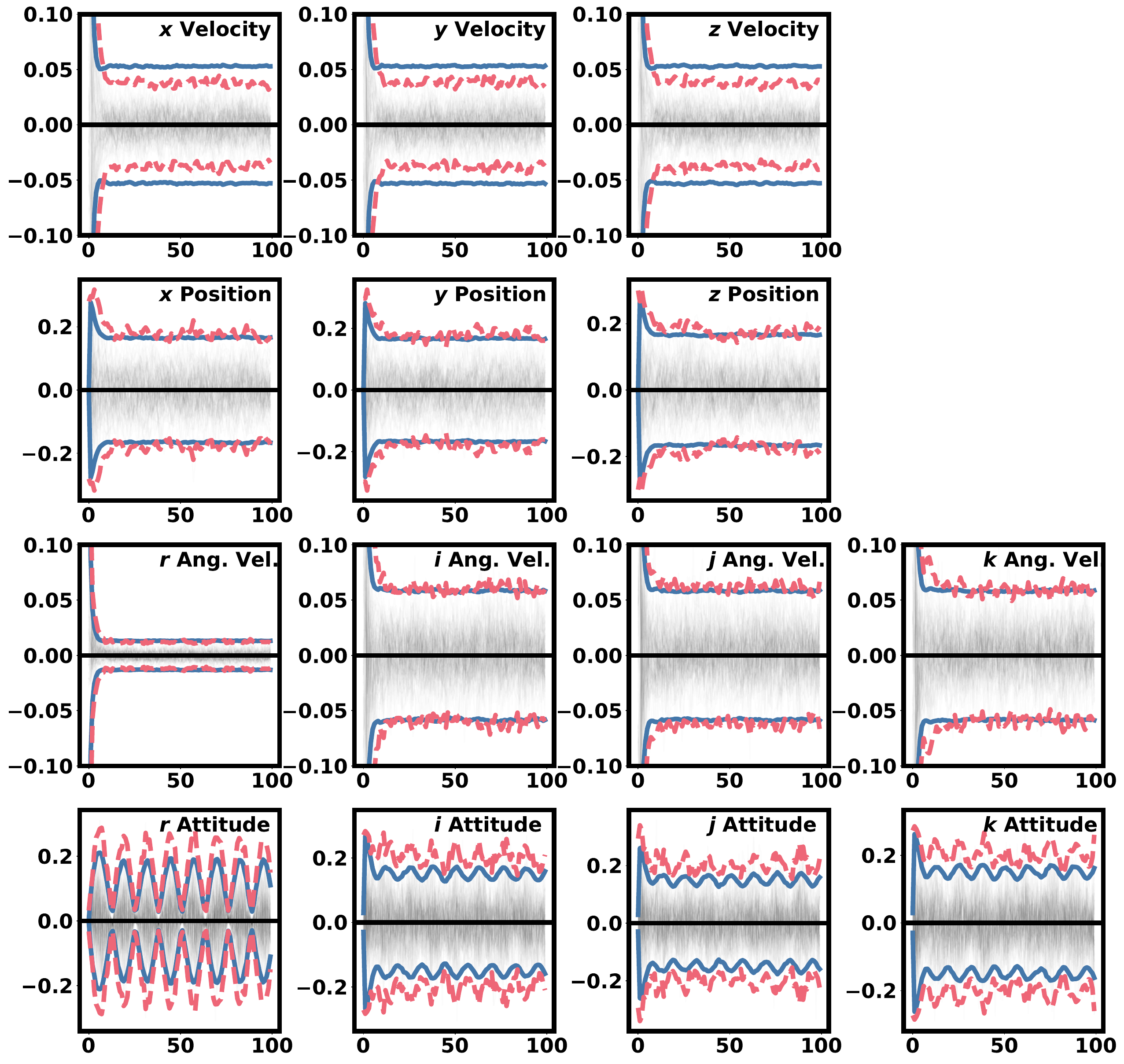}
    \begin{tikzpicture}
    \begin{axis}[%
    hide axis,
    xmin=10,
    xmax=50,
    ymin=0,
    ymax=0.4,
    every axis plot/.append style={line width=2pt, mark size=3.5pt},
    legend style={draw=white!15!black,legend cell align=left,legend columns=2}
    ]
    \addlegendimage{gray,mark=none, style={line width=0.5pt}}
    \addlegendentry{Monte Carlo runs};
    \addlegendimage{tolblue,mark=none}
    \addlegendentry{EnDKF predicted 3-$\sigma$ error};
    \addlegendimage{tolred,dashed,mark=none}
    \addlegendentry{EnDKF actual 3-$\sigma$ error};
    \end{axis}
    \end{tikzpicture}
    \caption{A plot of the error for the synthetic constant velocity example over all $14$ variables. The $x$-axes represent the time units, the $y$-axes represent error, the thin gray lines represent the errors from the Monte Carlo iterations, the blue lines represent the errors predicted by the EnDKF, as three standard deviations (3-$\sigma$) from the truth, while the dashed red lines represent the actual errors of the mean estimate, as three standard deviations from the truth.}
    \label{fig:synth-filter-results}
\end{figure}

\begin{figure}
    \centering
\includegraphics[width=0.6\linewidth]{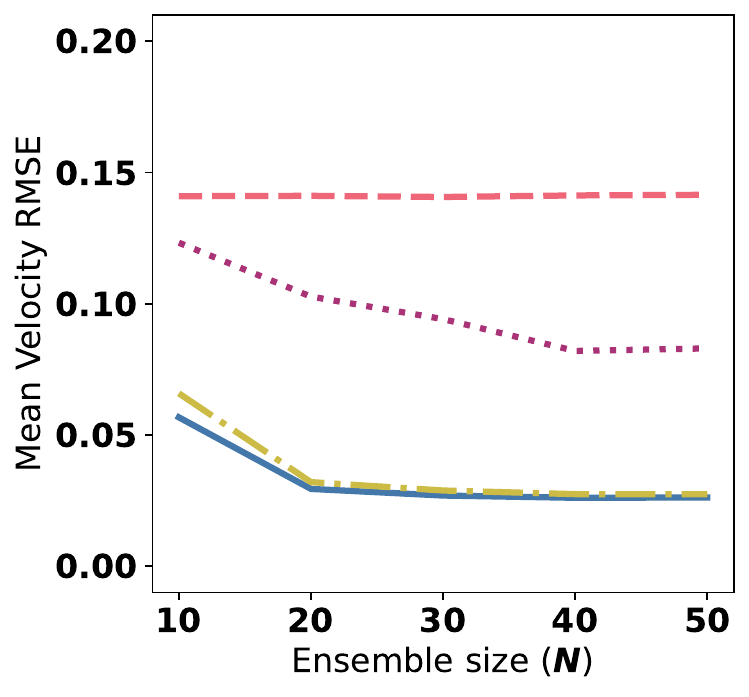}\\
\includegraphics[width=0.6\linewidth]{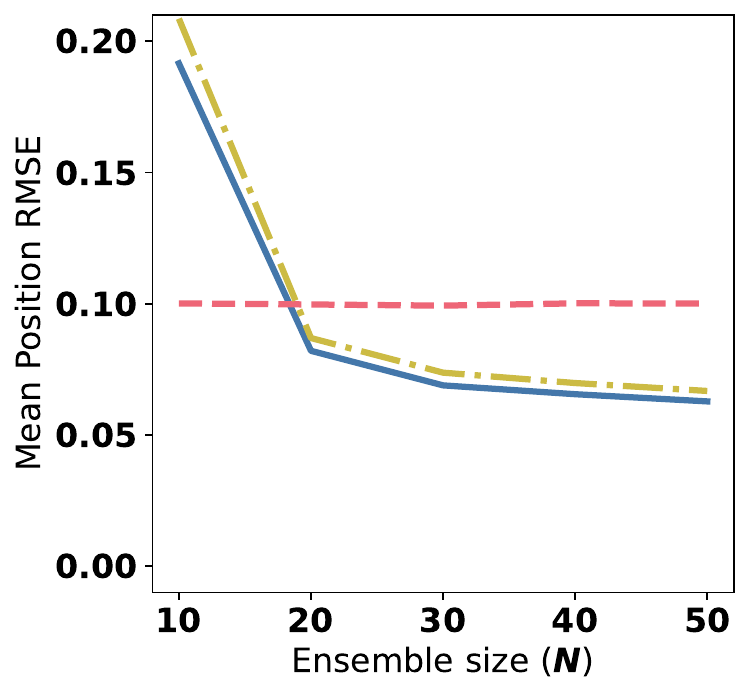}
    \begin{tikzpicture}
    \begin{axis}[%
    hide axis,
    xmin=10,
    xmax=50,
    ymin=0,
    ymax=0.4,
    every axis plot/.append style={line width=2pt, mark size=3.5pt},
    legend style={draw=white!15!black,legend cell align=left,legend columns=2}
    ]
    \addlegendimage{tolblue,mark=none}
    \addlegendentry{EnDKF};
    \addlegendimage{tolyellow,dashdotted, mark=none}
    \addlegendentry{EnDKF (no Adam)};
    \addlegendimage{tolpurple,dotted, mark=none}
    \addlegendentry{BPF};
    \addlegendimage{tolred,dashed,mark=none}
    \addlegendentry{Measurement};
    \end{axis}
\end{tikzpicture}
    \caption{Velocity and position RMSE for the synthetic data example. The $x$ axis represents the ensemble size of the EnDKF and the $y$ axis represents the mean RMSE. The solid blue line represents the error using the EnDKF, the dash-dotted yellow line represents the EnDKF with no Adam optimization performed, the dotted purple line represents the BPF, and the red line represents the error from only using the measurements.}
    \label{fig:pos-vel-rmse}
\end{figure}

\begin{figure}
    \centering
\includegraphics[width=0.6\linewidth]{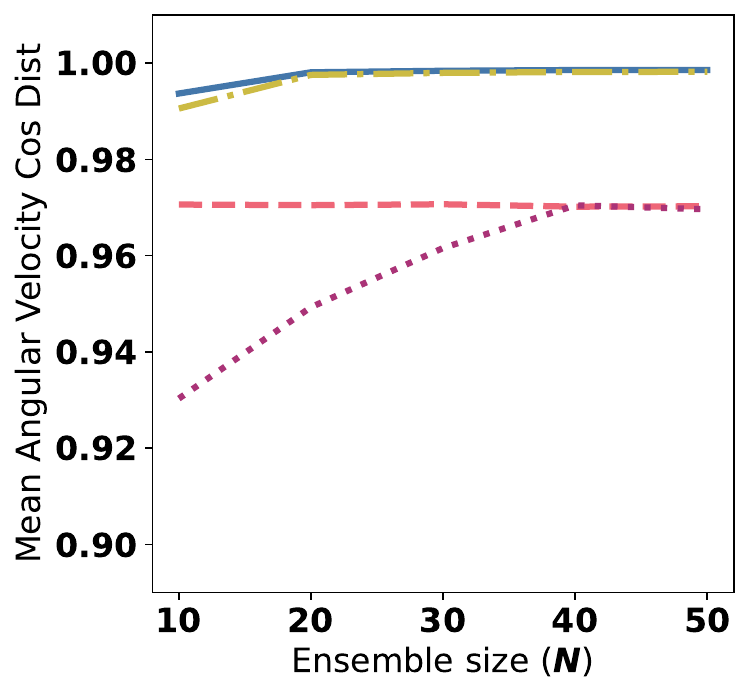}\\
\includegraphics[width=0.6\linewidth]{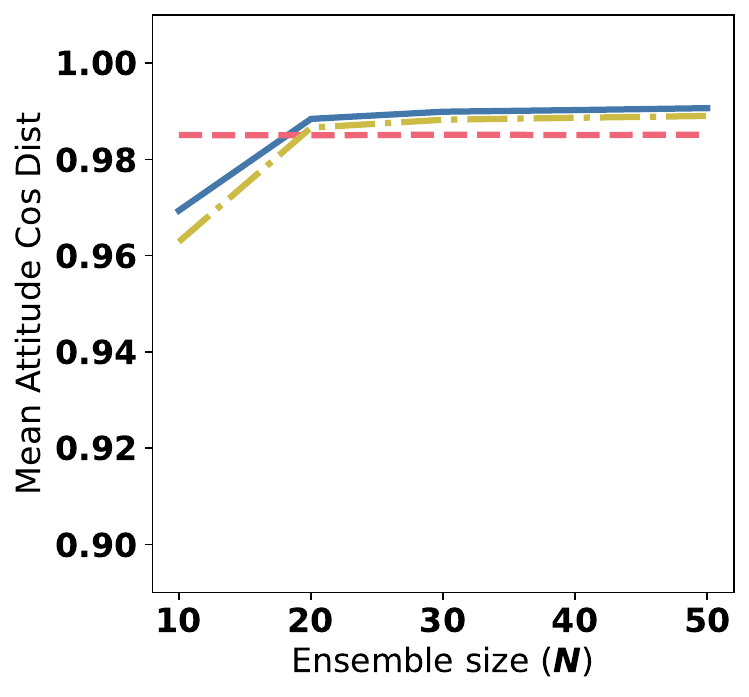}
    \begin{tikzpicture}
    \begin{axis}[%
    hide axis,
    xmin=10,
    xmax=50,
    ymin=0,
    ymax=0.4,
    every axis plot/.append style={line width=2pt, mark size=3.5pt},
    legend style={draw=white!15!black,legend cell align=left,legend columns=2}
    ]
    \addlegendimage{tolblue,mark=none}
    \addlegendentry{EnDKF};
    \addlegendimage{tolyellow,dashdotted, mark=none}
    \addlegendentry{EnDKF (no Adam)};
    \addlegendimage{tolpurple,dotted, mark=none}
    \addlegendentry{BPF};
    \addlegendimage{tolred,dashed,mark=none}
    \addlegendentry{Measurement};
    \end{axis}
\end{tikzpicture}
    \caption{Angular velocity and attitude cosine distance for the synthetic data example. The $x$ axis represents the ensemble size of the EnDKF and the $y$ axis represents the mean cosine distance. The solid blue line represents the error using the EnDKF, the dash-dotted yellow line represents the EnDKF with no Adam optimization performed, the dotted purple line represents the BPF, and the red line represents the error from only using the measurements.}
    \label{fig:pos-vel-cos}
\end{figure}

The numerical experiments presented herein serve two purposes: first, to validate the new ensemble Kalman filtering framework presented on the constant velocity model, and second, to apply this to object tracking with measurements obtained from a state-of-the-art foundations pose model, FoundationPose~\cite{wen2024foundationpose}.

\subsection{Constant Velocity Constant Rotation}

We begin through a simple validation of the EnDKF methodology presented herein without the use of a computer vision model.
We assume that there exists some object about which we want to have a measure of uncertainty. We assume that this object's state evolves in time through a noise-less (zero covarinace and concentration) constant velocity constant angular velocity model~\cref{eq:constant-velocity-model}.

We compare three filters: the i) EnDKF for the constant velocity model~\cref{eq:EnDKF-C} with Adam for solving the stochastic optimization problem, ii) the same algorithm except with only the naive Kalman gain computed as the initial condition and no stochastic optimization, and iii) the bootstrap particle filter~\cite{reich2015probabilistic}.

Take the true state at the initial time unit $k=0$, to be,
\begin{equation}
    T_{p,0} = \begin{bmatrix}
        0&0&0
    \end{bmatrix}^T,
\end{equation}
and the constant velocity at all time units to be,
\begin{equation}
    T_{v} = \begin{bmatrix}
        0.1&0.1&0.1
    \end{bmatrix}^T.
\end{equation}
We assume that object is initially not rotated with relation to some reference frame, meaning that, at the initial time unit $k=0$, the attitude of the object is set to,
\begin{equation}
    T_{\omega,0} = \begin{bmatrix}
        1&0&0&0
    \end{bmatrix}^T,
\end{equation}
which is a unit quaternion.
The constant rotation at all time we represent by,
\begin{equation}
    T_{\dot{\omega}} = \begin{bmatrix}
        \sqrt{47/50}& \sqrt{1/50}& \sqrt{1/50}& \sqrt{1/50}
    \end{bmatrix}^T,
\end{equation}
which again is a unit quaternion.

where the position measurement error is unbiased Gaussian with covariance,
\begin{equation}
    R = 10^{-2},
\end{equation}
and the attitude error is Fisher-von Mises Langevin distributed with mean direction at the truth, and concentration,
\begin{equation}
    \rho = 10^2.
\end{equation}
For the EnDKF ensembles of perturbed observations were computed using~\cref{eq:Rn-perturbed} for the position, and~\cref{eq:Snm1-perturbed} for the velocity.

Our estimate of the position and attitude at the initial time is determined by the first measurement,
\begin{equation}
    X_{p, 0} = Y_{p, 0},\,\, X_{\omega, 0} = Y_{\omega, 0},
\end{equation}
while our initial velocity and initial angular velocity are determined by crude approximations from the first two measurements,
\begin{equation}\label{eq:initial-velocity}
    X_{v, 0} = Y_{p, 1} - Y_{p, 0},\,\, X_{\dot\omega, 0} = Y_{\omega, 1} Y_{\omega, 0}^{-1},
\end{equation}
where the multiplication and inversion in the last equation is that of quaternion multiplication and inversion.
The initial ensembles are generated with 

The process noise for position $p$ and velocity $v$ we define by the following covariances,
\begin{equation}
    Q_v = 10^{-4} I_3, \,\, Q_p = 10^{-8} I_3,
\end{equation}
and the concentration of the angular velocity $\dot\omega$ and attitude $\omega$, we define to be, 
\begin{equation}
    \text{\Qoppa}_{\dot\omega} = 10^4, \,\, \text{\Qoppa}_\omega = 10^6.
\end{equation}

For the velocity and position error metric we compute the mean root-mean-squared error,
\begin{equation}\label{eq:RMSE}
\begin{gathered}
    \operatorname{RMSE}_{p/v} = \frac{1}{M}\sum_{m=1}^M\sqrt{\frac{1}{3T}\sum_{i,k}(p/v^\text{true} - \widetilde{p/v})^2_{i,k}}\\
    \widetilde{p/v} = \frac{1}{N}\sum_{j=1}^N [E_{p/v}]_j
\end{gathered}
\end{equation}
where $T$ is the number of time steps, $M$ is the number of Monte Carlo runs, $i$ is the space index representing the $x$, $y$, and $z$ coordinates, $k$ is the time index, and $m$ is the Monte Carlo index. A lower $\operatorname{RMSE}$ is better.

For the angular velocity and attitude error we use the cosine distance,
\begin{equation}\label{eq:cos-dist}
    \cos_{\dot\omega/\omega} = \frac{1}{MN}\sum_{m=1}^M \sum_{j=1}^N (\dot\omega/\omega)^\text{true}\cdot (\dot\omega/\omega)
\end{equation}
where $\cdot$ is the dot product, and all other variables are the same as before. A $\cos$ distance closer to $1$ is better.

For the stochastic optimization, 25 steps of Adam with a step-size/learning-rate of $\alpha = 10^{-1}$ are performed.

The true model is run for 100 arbitrary time units over 100 Monte Carlo iterations with an ensemble size ranging in the set $N\in\{10,20,30,40,50\}$. Each Monte Carlo run has the same truth but different realizations of the measurements.

We first look at the plots of the raw errors for the EnDKF for the $14$ variables in~\cref{fig:synth-filter-results} for an ensemble size of $N=50$. As can be seen, other than at burn-in time, the filter is conservative in velocity, while the prediction of the error in position and angular velocity are very precise. The prediction of the attitude error is not conservative, meaning that the error is larger than predicted.
Note that there is no bias at all in the predictions, meaning that the filter is unbiased for this scenario, even when it is not really linear.

Next we look at the velocity and position RMSE~\cref{eq:RMSE} of the EnDKF as compared to using only measurements directly. For the measurement velocity, it is estimated using a first order forward finite difference just as in~\cref{eq:initial-velocity}. As can be seen in~\cref{fig:pos-vel-rmse}, the EnDKF velocity RMSE for all ensemble sizes, except $N=10$, is significantly smaller than that estimated by the measurements with a $81\%$ reduction in error in the case of $N=50$. For the position error the discrepancy is not quite that high, with the most significant in error reduction being $37\%$ in the case of $N=50$. We note that the EnDKF with no Adam optimization performs worse, but not significantly worse than the EnDKF with Adam optimization. This means that while the Adam optimized version is indeed more optimal, it is not significantly more optimal.
We also note that the BPF failed to predict the position with any accuracy for the given number of particles, which is not surprising as the number of particles required for the BPF to function is usually exponential in the dimension. 
More surprising is the result that the BPF is capable predicting the velocity better than the measurements, which, given the lack of accuracy in the position prediction is likely accidental.

Finally we look at the angular velocity and attitude cosine distance error~\cref{eq:cos-dist} of the EnDKF as compared to using only measurements directly. For the measurement angular velocity, it is estimated using a~\cref{eq:initial-velocity}. As can be seen in~\cref{fig:pos-vel-cos}, the EnDKF angular velocity cosine distance is significantly closer to one than that of purely using the measurements, with a $95\%$ reduction in error in the case of $N=50$. The attitude cosine distance reduction is more modest, with a $37\%$ reduction in error in the case of $N=50$. The results with the EnDKF with no Adam and the BPF are similar to the position and velocity cases.

These results conclusively show that the EnDKF is capable of performing linear tracking in the case of a twin-experiment where the object is indeed moving at a constant velocity with a constant angular velocity applied to its attitude.

\subsection{Head Tracking in Digital Twin Environment}

\begin{figure}
    \centering
    \includegraphics[trim={0 0 12cm 0},clip,width=0.8\linewidth]{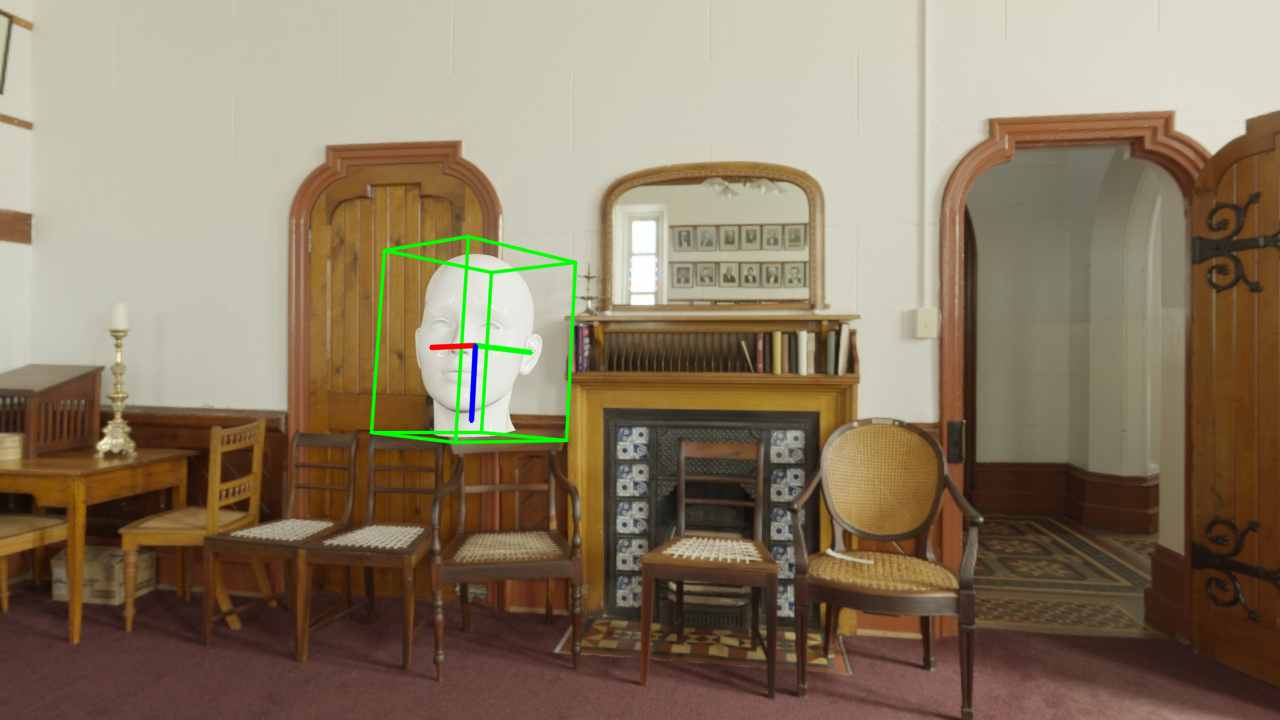}
    \caption{A measurement of the attitude of the simulated head juxtaposed to the frame taken by the simulated camera in a digital twin environment.}
    \label{fig:head-in-virtual-environment}
\end{figure}

\begin{table}[t]
    \centering
\begin{tabularx}{0.9\linewidth}{Xlll}\toprule
Error & Measurement & EnDKF & Impr.\\\toprule
$\operatorname{RMSE}_v$ & 6.39791e-05  cm/f & 1.78784e-05 cm/f & 72.06\%\\
$\operatorname{RMSE}_p$ & 1.03633e-03  cm & 9.78717e-04 cm & 5.56\%\\
$1 - \cos_{\dot\omega}$ & 1.29023e-06 & 3.52336e-08 & 97.27\% \\
$1 - \cos_{\omega}$ & 3.00505e-05 & 2.98162e-05 & 0.78\% \\\hline\\
\end{tabularx}
\caption{Ideal head measurements and filtering with no additional bias. The errors for position are in centimeters and for velocity are in centimeters per frame.}\label{tab:head-perfect}
\vspace{-20pt}
\end{table}

\begin{table}[t]
    \centering
\begin{tabularx}{0.9\linewidth}{Xlll}\toprule
Error & Measurement & EnDKF & Impr.\\\toprule
$\operatorname{RMSE}_v$ & 1.44964e-04 cm/f & 7.92701e-05  cm/f & 45.32\%\\
$\operatorname{RMSE}_p$ & 6.66132e-03 cm & 6.19365e-03 cm & 7.02\%\\
$1 - \cos_{\dot\omega}$ & 2.39570e-05 & 7.98386e-06 & 66.67\% \\
$1 - \cos_{\omega}$ & 1.16130e-03 & 9.66479e-04 & 16.78\% \\\hline\\
\end{tabularx}
\caption{Head measurements and filtering with bias introduced by vertex displacement, and a different head model fed into the FoundationPose model. The errors for position are in centimeters and for velocity are in centimeters per frame.}\label{tab:head-bad-all}
\vspace{-20pt}
\end{table}

For the second experiment, we construct a digital twin of a pose-tracking scenario involving a low-feature, uniform-albedo human head in an indoor environment. 
The scene is illuminated using a high-dynamic-range image (HDRI) environment map of a church meeting room~\cite{polyhaven_church_meeting_room}, while the target object is a computer-aided design model of a human-like head~\cite{thingiverse_head_979818}, whose pose we seek to estimate. 
The digital twin is implemented in the Blender 3D content-creation suite via BlenderProc~\cite{denninger2019blenderproc,denninger2023blenderproc2}, which is also used to generate the synthetic dataset. 
Virtual image capture is configured to match the intrinsic parameters of a real-world camera (an Orbbec Femto Mega), with calibration $f_x = 748.49$, $f_y = 748.20$, $c_x = 623.06$, $c_y = 342.77$ at a resolution of $1280\times720$.
The camera is positioned $1.0$ meter from the head along the Y-axis and oriented toward the origin.
FoundationPose~\cite{wen2024foundationpose} is employed to generate pose estimates over time, with the virtual camera operating at $10$ frames per second. 
A visualization of the environment, along with an example attitude estimate, is shown in~\cref{fig:head-in-virtual-environment}.

As this setup constitutes a digital twin, ground-truth pose information is available, enabling direct comparison with estimated results in the same manner as in the previous experiment. 
The head is rotated with a constant yaw angle of $10.8$ minutes per frame, corresponding to a quaternion that is approximately~\cite{henderson1977euler,bernardes2022quaternion},
\begin{equation}
    \dot\omega = \begin{bmatrix}
        9.99 \cdot 10^{-1} & 0 & -1.57 \cdot 10^{-3} & 0
    \end{bmatrix}^T,
\end{equation}
while its position remains fixed throughout the sequence.

We first look a the ideal scenario, when the head that is rendered is the head that is exactly known by the FoundationPose algorithm.
The process we define to be for position $p$ and velocity $v$ we define by the following covariances,
\begin{equation}
    Q_v = 10^{-8} I_3, \,\, Q_p = 10^{-8} I_3,
\end{equation}
 the concentration of the angular velocity $\dot\omega$ and attitude $\omega$, we define to be, 
\begin{equation}
    \text{\Qoppa}_{\dot\omega} = 10^8, \,\, \text{\Qoppa}_\omega = 10^8,
\end{equation}
and we assume that the measurement noise has the covariance and concentration of,
\begin{equation}
    R = 1e-6 I_3, \,\, \rho = 1e5.
\end{equation}
We now only compare the EnDKF~\cref{eq:EnDKF-C} with the number of ensemble members set to a constant $N=500$, with a constant velocity model~\cref{eq:constant-velocity-model}, and see if it can outperform only using the measurements.
The Adam optimization uses a step size of $\alpha = 10^{-2}$ and $250$ steps of the algorithm.
For the velocity and position we again look at the RMSE~\cref{eq:RMSE}, and for the angular velocity and attitude, we look at one minus the cosine distance~\cref{eq:cos-dist}, in order to see the improvement in the error. The results can be seen in~\cref{tab:head-perfect}. As can be seen the EnDKF is also better than the measurements at predicting the velocity and the angular velocity, but the improvement in error is marginal for the more important states of position and attitude.
This means that there is no significant benefit gained from using the EnDMF in this `perfect' scenario.

In a real tracking scenario, it is not possible to know the 3D model of the object that we are attempting to track, therefore, for our third experiment we introduce a more real-world scenario: we assume that the computer-aided-design file head which we provide to the FoundationPose model is not perfect. For this work a new head was generated using a generative model~\cite{xiang2025nativecompactstructuredlatents}.
Additionally the rendered head~\cref{fig:head-in-virtual-environment} is perturbed through vertex displacement~\cite{lee2000displaced,cook1984shade} (slight perturbations to the vertex indices as described in the appendix),
to create a much more difficult surface to measure.

For vertex perturbations, mesh coordinates are represented in meters, while
the displacement magnitude is specified in millimeters. Take $w_i~\mathcal{N}(0, 1)$ to standard Gaussian noise for the $i$th vertex. Each \(w_i\) is clipped to the interval
\([-2,2]\), producing the initial noise value. We then perform
three rounds of 1-ring neighborhood smoothing~\cite{taubin1995signal}:
\begin{equation}
    s_i \leftarrow 0.55 s_i + 0.45 \frac{1}{\lvert N_i \rvert} \sum_{j\in N_i} s_j,
\end{equation}
where \(N_i\) is the 1-ring neighborhood of vertex \(i\).
The smoothed value is mixed with the original clipped sample and converted to a physical displacement by
\begin{equation}
    m_i = 0.65 w_i + 0.35 s_i,\,\, d_i = \alpha m_i,
\end{equation}
where \(\alpha = 0.75\,\mathrm{mm}\). The perturbed vertex is then,
\begin{equation}
    \mathbf{v}'_i = \mathbf{v}_i + d_i \mathbf{n}_i,
\end{equation}
where \(\mathbf{v}_i\) is the original vertex position and \(\mathbf{n}_i\)
is the unit surface normal.

In the EnDKF, we model these corruptions to the underlying measurements in a constant bias model as in~\cref{eq:measurement-pos-att-bias}. As the model is now more corrupted we increase the amount of process noise in this scenario.
The process we define to be for position $p$ and velocity $v$ we define by the following covariances,
\begin{equation}
    Q_v = 10^{-7} I_3, \,\, Q_p = 10^{-7} I_3,
\end{equation}
 the concentration of the angular velocity $\dot\omega$ and attitude $\omega$, we define to be, 
\begin{equation}
    \text{\Qoppa}_{\dot\omega} = 10^7, \,\, \text{\Qoppa}_\omega = 10^7,
\end{equation}
with the process noise for the bias~\cref{eq:constant-bias-model}, having covariance and concentration of,
\begin{equation}
     Q_b = 10^{-8} I_3, \,\, \text{\Qoppa}_\beta = 10^8,
\end{equation}
and we assume that the measurement noise has the covariance and concentration of,
\begin{equation}
    R = 1e-6 I_3, \,\, \rho = 1e5.
\end{equation}
The initial bias in position is assumed to be zero and the initial bias in attitude is assumed to be the unit quaternion with real part $1$.

The results of this experiment can be seen in~\cref{tab:head-bad-all}.
As can be seen, the bias model accounts for error in the position better than in the ideal head experiment with a $7\&$ improvement in the error, and for a large $16\%$ improvement in error for the attitude. The velocity and angular velocity error improvements are gain high, but not as high as they were in the ideal head scenario.
These results show that a simple constant bias model can account for large complex non-linear non-normal bias introduced by an imperfect computer vision model.

\section{Conclusions}
\label{sec:conclusions}
This paper addressed the limitations of vision-based systems in accurately estimating pose, where measurement errors can persist and become critical in precision-sensitive applications such as brain surgery. 
The proposed ensemble directional Kalman filter (EnDKF) introduces a hybrid Euclidean-directional representation for pose uncertainty, and linear manipulation thereof.

Experimental results demonstrate that the proposed framework achieves reliable and consistent pose estimation under noisy and uncertain conditions, on both synthetically generated data and measurement data collected from a digital twin.
Despite these promising results, several limitations remain. 
The current evaluation relies in part on synthetic data and simplified models, which may not fully capture the variability and complexity of real-world scenarios. 
Furthermore, the stochastic optimization of the Kalman gain introduces additional computational overhead and requires careful tuning.

Future work will focus on validating the proposed method using real measurement data, including experiments involving real-world subjects rather than simulated models.
In addition, alternative optimization strategies will be explored to further improve the speed and accuracy of the Kalman gain estimation.

\section*{Acknowledgment}
A large language model interface, Perplixity.ai was used for improving language and clarity during the editing process set on ``best model'' mode.

\bibliographystyle{IEEEtran}
\bibliography{biblio}

\begin{IEEEbiography}[{\includegraphics[width=1in,height=1.25in,clip,keepaspectratio]{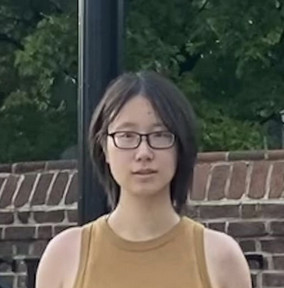}}]{Tianlu Lu} received the B.S. degrees in Computer Science and Mathematics from Frostburg State University in 2022 and the M.P.S. degree in Data Science and Analytics from the University of Maryland, College Park, in 2023. She is currently pursuing the Ph.D. degree in Computer Science at the University of Hawaii at Manoa under the supervision of Dr. Andrey A. Popov. Her research focuses on data assimilation.
\end{IEEEbiography}

\begin{IEEEbiography}[{\includegraphics[width=1in,height=1.25in,clip,keepaspectratio]{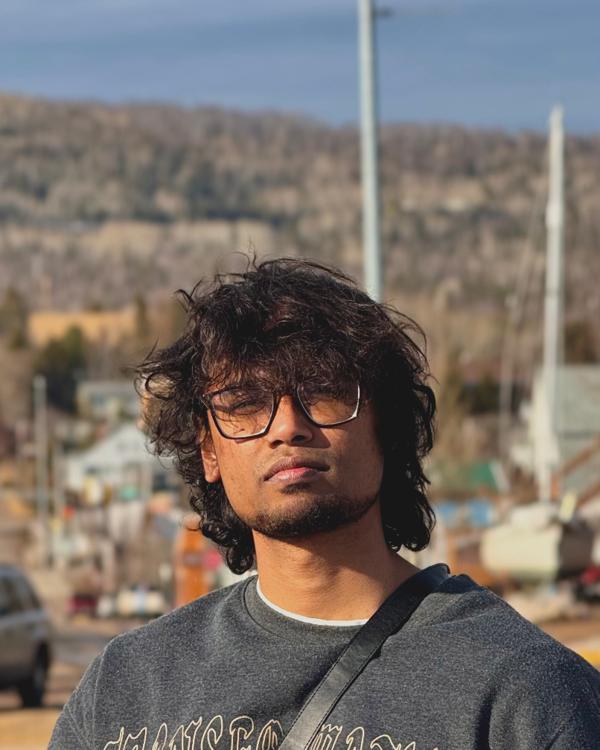}}]{Asif Sijan}
is a Ph.D. student in the Department of Information and Computer Sciences at the University of Hawai‘i at Mānoa. He received his B.S. from American International University-Bangladesh (AIUB) and his M.S. from the University of Minnesota Duluth. His research focuses on computer vision and 3D scene understanding, including multi-view geometry, geometric reconstruction, 3D Gaussian Splatting (3DGS), and immersive visual computing.

\end{IEEEbiography}

\begin{IEEEbiography}[{\includegraphics[width=1in,height=1.25in,clip,keepaspectratio]{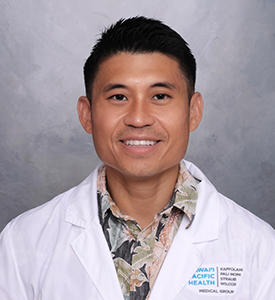}}]{Thomas Noh}, MD, is a neurosurgeon and serves as Chief Business and Strategy Officer at Excel Health which he co-founded in Honolulu, Hawai'i. He is a Clinical Assistant Professor at the University of Hawai'i John A. Burns School of Medicine and an affiliated faculty member in the Department of Information and Computer Sciences at the University of Hawai'i at Manoa.
Dr. Noh completed advanced neurosurgical fellowship training in image-guided and minimally-invasive surgery at Harvard, Houston Methodist and the University of Michigan. His clinical and research interests focus on image-guided surgery, endoscopic spine techniques, and the integration of artificial intelligence to enhance surgical precision and outcomes.
He has been a pioneer in advancing outpatient spine surgery in Hawai'i, performing the state’s first endoscopic spine procedures and developing intraoperative imaging protocols. His work includes NIH-funded research and multidisciplinary collaborations spanning neurosurgery, engineering, and computer science.
\end{IEEEbiography}

\begin{IEEEbiography}[{\includegraphics[width=1in,height=1.25in,clip,keepaspectratio]{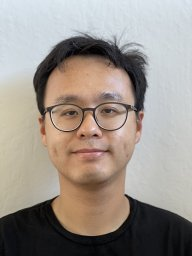}}]{Huaijin Chen} is an Assistant Professor of Computer Science at the University of Hawai'i at Manoa, where he directs the Computational Imaging and Robotic Perception (CIRP) Lab. He earned his Ph.D. in Electrical and Computer Engineering from Rice University in 2019 and a B.S. in Imaging Science from the Rochester Institute of Technology. Before academia, he worked at companies like Vayu Robotics, NVIDIA, and IBM. He has authored over 15 peer-reviewed papers in top-tier venues such as CVPR, ICCP, and Optics Express, and holds four U.S. patents. Dr. Chen has served on program committees for ICCP and IJCAI, and is an active reviewer for major journals and conferences such as IEEE TPAMI, IEEE TIP, Optics Express, Optics Letters, CVPR, ICCV, ECCV, and ICLR. His awards include the Outstanding Reviewer Award at ICCV 2021, the Best Poster Award at ICCP 2019, and the Texas Instruments Distinguished Graduate Student Fellowship.
\end{IEEEbiography}
 
\begin{IEEEbiography}[{\includegraphics[width=1in,height=1.25in,clip,keepaspectratio]{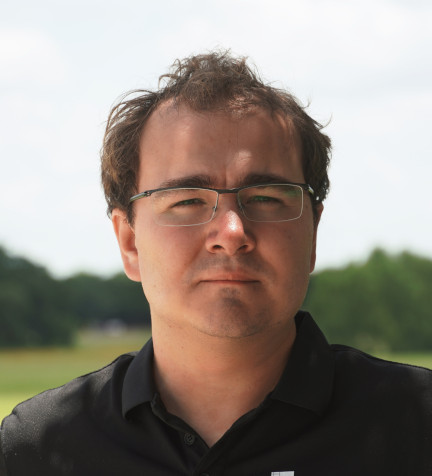}}]{Andrey A. Popov}
received his B.S. in Mathematics from RPI, his Ph.D. in Computer Science from Virginia Tech, and was a postdoctoral fellow at the Oden Institute for Computational Engineering and Sciences at the University of Texas at Austin.
He is currently an Assistant Professor at the Department of Information \& Computer Sciences at the University of Hawai'i at Manoa.
He has authored over 44 peer-reviewed papers in journals and conferences including in SIAM SISC and FUSION.
In 2024 he was awarded the Jean-Pierre Le Cadre prize for the best paper at the International Conference on Information Fusion (FUSION).
His primary research focuses on fusing theory-driven and data-driven methods for scientific applications, data assimilation, and aerospace problems. His other research interests include directional statistics, and geographic information systems.
\end{IEEEbiography}

\end{document}